\documentclass[preprint,12pt]{elsarticle}

\usepackage{amssymb}
\usepackage[utf8]{inputenc}
\usepackage{array}
\usepackage{multirow}
\usepackage{booktabs}
\usepackage{graphicx}
\usepackage{float}
\usepackage{amsmath}
\usepackage{tcolorbox}
\usepackage{algorithm}
\usepackage{algorithmic}
\usepackage[margin=1in]{geometry}
\usepackage{multicol}

\usepackage{subcaption}

\journal{iScience of the Cell Press}

\begin{document}

\begin{frontmatter}



% \title{From Knowledge Generation to Knowledge Verification: A Deep Dive into the Generative Capabilities of ChatGPT}

\title{From Knowledge Generation to Knowledge Verification: Examining the BioMedical Generative Capabilities of ChatGPT}


\author[inst1,inst3]{Ahmed Abdeen Hamed*}

\affiliation[inst1]{organization={MGEN -- College of Engineering, Northeastern University Miami},%Department and Organization
            % addressline={Address One}, 
            city={Miami},
            postcode={33127}, 
            state={FL},
            country={United States}}

\author[inst4]{Alessandro Crimi}
\author[inst5]{Magdalena M Misiak}
\author[inst2]{Byung Suk Lee}

\affiliation[inst2]{organization={Department of Computer Science, University of Vermont},%Department and Organization
            % addressline={Address Two}, 
            city={Burlington},
            postcode={05405}, 
            state={VT},
            country={United States}}
            
\affiliation[inst3]{organization={CASCI Laboratory, Binghamtom University},%Department and Organization
            % addressline={Address Two}, 
            city={Binghamton},
            state={NY},            
            postcode={13902}, 
            country={United States}}

\affiliation[inst4]{organization={AGH, University of Krakow, Faculty of Informatics},%Department and Organization
            % addressline={Address Two}, 
            city={Krakow},          
            postcode={30-059}, 
            country={Poland}}
            
\affiliation[inst5]{organization={Department of Physiology and Biophysics, Howard University},%Department and Organization
            % addressline={Address Two}, 
            city={Washington},
            state={DC},            
            postcode={20059}, 
            country={United States}}

\begin{abstract}
The generative capabilities of LLM models offer opportunities for accelerating tasks but raise concerns about the authenticity of the knowledge they produce. To address these concerns, we present a computational approach that evaluates the factual accuracy of biomedical knowledge generated by an LLM. Our approach consists of two processes: generating disease-centric associations and verifying these associations using the semantic framework of biomedical ontologies. Using ChatGPT, as the selected LLM, we designed prompt-engineering processes to establish linkages between diseases and related drugs, symptoms, and genes, and assessed consistency across multiple ChatGPT models (e.g., GPT-turbo, GPT-4, etc.) Experimental results demonstrate high accuracy in identifying disease terms (88\%--97\%), drug names (90\%--91\%), and genetic information (88\%--98\%). However, symptom term identification was notably lower (49\%--61\%), due to the informal and verbose nature of symptom descriptions, which hindered effective semantic matching with the formal language of specialized ontologies. Verification of associations reveals literature coverage rates of 89\%--91\% for disease-drug and disease-gene pairs, while symptom-related associations exhibit lower coverage (49\%--62\%). Despite the high accuracy in identifying term names, our observations revealed that the generated term IDs were invalid and, in some cases, redundant. This issue may impede knowledge integration across multiple ontologies, as term IDs are essential for establishing cross-references. Overall, these findings suggest that GenAI tools (e.g., ChatGPT) can generate more reliable biomedical knowledge if extra care is taken. Employing strategies such as Retrieval Augmented Generation (RAG) and incorporating knowledge from literature or ontology terms, as supervised in-context prompting, may further enhance the accuracy and reliability of GenAI-generated biomedical knowledge. Further work is needed to investigate these directions.
\end{abstract}

\begin{highlights}
\item Evaluating the capabilities of LLMs, exemplified by ChatGPT, to generate disease-centric biomedical associations.
\item Validating the correctness of biomedical named entity terms (including diseases, drugs, symptoms, and genetics) using specialized biomedical ontologies.
\item Assessing the reliability of structured associations by cross-referencing them with biomedical literature databases (e.g., PubMed), ensuring alignment with established scientific evidence.
\item Examining the consistency of LLMs in reproducing their generated associations within simulated biomedical abstracts across various ChatGPT models.
\item Demonstrating how improved validation strategies can enhance biomedical knowledge generation and knowledge integration, potentially informing better the biomedical research.
\end{highlights}

\end{frontmatter}

\newpage
\section*{Graphical Abstract}
\begin{figure}[htbp]
   \centering
   \includegraphics[scale=.12]{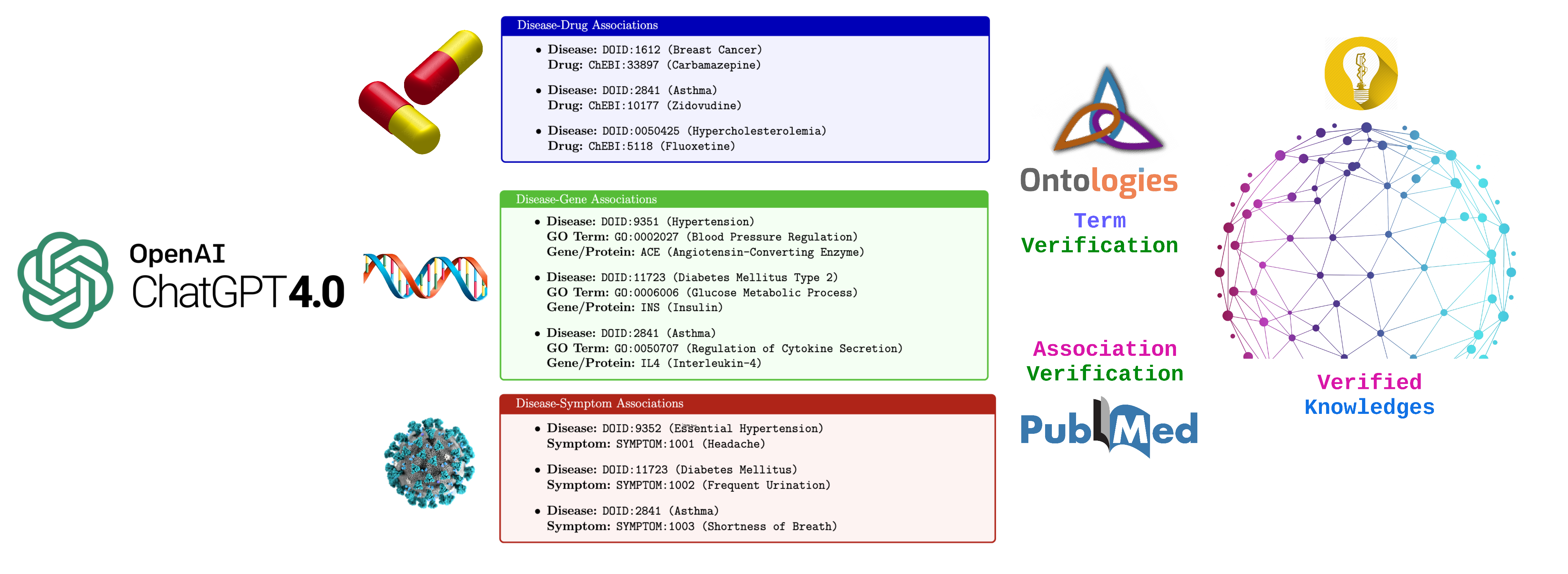}
   \caption{Graphical Abstract of the Knowledge Generation and Verification Tasks Performed.}
    % \label{fig:graphical_abstract}
\end{figure}

% {
% \centering
% \includegraphics[scale=.12]{iScience-Fact-Checking-GA.png}\\    
% Knowledge generation and verification tasks performed.
% }

\newpage

\section{Background and Significance} 
Large language models (LLMs) possess impressive generative capabilities, presenting opportunities to accelerate various tasks while also raising concerns about the reliability of the knowledge they generate. Concerned with the growing threat to scientific authenticity since the emergence of ChatGPT \cite{chatgpt2025}, our previous perspective laid out a research agenda focused on safeguarding authenticity \cite{hamed2024safeguarding}. This agenda addressed two key issues: combating fake science and verifying the factuality of AI-generated content. The first issue, fighting fake science, has become increasingly urgent as ChatGPT's generative capabilities can produce convincing but fabricated scientific articles. In our earlier work, we introduced the xFakeSci algorithm, a machine learning approach that distinguished fake/AI-generated articles from real publications with high precision \cite{hamed2024detection}. This development underscores the need for robust mechanisms to detect and filter out misleading content. The second issue, which is the focus of this paper, is the verification of the factuality of content generated by ChatGPT. Here, we extend our previous efforts by developing algorithmic approaches to assess and verify biomedical associations. Our experiments leverage relevant biomedical ontologies and PubMed abstracts to systematically evaluate the factual accuracy of the generated content.

\section{Introduction}
In an era marked by the rapid adoption of generative AI tools, fact-checking and knowledge verification are essential safeguards against hallucination and misinformation. This review traces the evolution of fact-checking, from its pre-ChatGPT foundations to its current role in addressing challenges posed by advanced generative models, especially in biomedical applications.

\subsection{Literature review}
Prior to the emergence of ChatGPT, research efforts focused on examining misinformation and developing fact-checking strategies. Several studies investigated the authenticity of information and explored fact-checking methods to mitigate misinformation risks~\cite{ciampaglia2015computational,luengo2020performance,nyhan2020taking,rodriguez2021debunking}. Surveys provided a comprehensive overview of automated fact-checking models and databases~\cite{zeng2021automated,guo2022survey}, while others employed Natural Language Processing techniques to verify news articles and social media content~\cite{lazarski2021using,oshikawa2018survey}. Additionally, machine learning approaches were applied to combat fake news, fake science, and fake social media posts~\cite{anusree2022factorfake,khalil2021detecting,zhou2019physiological,vo2019learning}. The urgency of these efforts was further underscored during the global pandemic, as misinformation raised significant health and public safety concerns~\cite{krause2020fact,luengo2020performance,abdeen2021fighting,siwakoti2021covid}.

The release of ChatGPT, alongside other generative AI and large language models, expanded research opportunities while intensifying concerns about misinformation. On one hand, these models have unlocked new scientific possibilities~\cite{koohi2023generative,degrave2023dissection,trabassi2024optimizing,wang2023applications}; on the other, they have raised issues regarding hallucinations and the lack of citations, which challenge scientific authenticity~\cite{tian2024opportunities,van2024chatgpt,barreto2023generative,giannakos2024promise}. In response, new fact-checking approaches emerged. For instance, one study addressed the verification of simulated medical abstracts by examining disease and gene names~\cite{Hamed_FC2004}, while another explored fact-checking solutions to mitigate risks associated with factuality in large language models~\cite{augenstein2024factuality}. Additional systems, such as LLM-Augmenter, have been designed to cross-verify content against external resources, and deep-learning classifiers have been used to check AI-generated radiology reports~\cite{peng2023check,mahmood2023fact}. Moreover, SelfCheckGPT has been developed to assess factuality on a sentence-by-sentence basis by ranking text chunks~\cite{manakul2023selfcheckgpt}.

Biomedical research has benefited from early models like BioBART, which assisted with Named Entity Recognition (NER), Entity Linking, and Question Answering tasks at a limited scale~\cite{yuan2022BioBART}. Following ChatGPT’s debut, studies began exploring its utility in biomedical question-answering~\cite{jin2023retrieve,hou2023answers}. Concurrently, researchers have investigated the use of large language models to generate knowledge directly or via Retrieval-Augmented Generation (RAG) methods. RAG integrates contextual prompts to enhance the freshness and accuracy of the generated information~\cite{huly2024old,jeong2023generative,arslan2024business,ng2024rag}. For example, one study employed ChatGPT as a decision support system for self-screening by embedding screening guidelines into hypothetical cases~\cite{khan10803468MedAI}. Another used RAG-based prompt engineering to extract structured representations of drug combinations from clinical trials~\cite{Hamed10803434MedAI}. Similar approaches have also improved PubMed’s retrieval capabilities~\cite{thomo2024pubmed}.

Beyond biomedical applications, large language models have been evaluated for their fact-checking abilities in news and multilingual settings. Comparative studies of ChatGPT, Bing AI CoPilot, and Gemini (formerly Bard) have highlighted both their potential and the continuing need for human oversight in news verification~\cite{chatgpt2025,copilot2025,gemini2025,caramancion2023news}. In the multilingual arena, research employing techniques such as zero-shot, chain-of-thought, and cross-lingual prompting has shown that languages with fewer resources may sometimes yield more accurate fact-checking results~\cite{singhal2024multilingual}. Additionally, studies have underscored the importance of developing guidelines for using AI in fact-checking news headlines~\cite{DeVerna_pnas_2322823121}.

\subsection{Goal and objectives}
The goal of our work is to test the generative capabilities of ChatGPT to generate disease-centric biomedical terms and associations and perform various verification process to assess the factuality of such associations. There are three objectives under this goal.

\noindent\textbf{Objective 1:} To perform semantic term verification using relevant biomedical ontologies for disease, drugs, symptoms, and genes. 

\noindent\textbf{Objective 2:} To perform automated verification of association using against the biomedical literature, specifically, PubMed abstracts. 

\noindent\textbf{Objective 3:} Assess ChatGPT consistency in generating knowledge using independent processes and various ChatGPT models: gpt-turbo, gpt-4o, gpt-4, and gpt-4o-mini.

\section{Experiment Tasks}
\textbf{Task 1 (Term correctness verification):} This task is to verify the biomedical terms that make up a generated association, where we use biomedical ontology, such as GO, DOID, ChEBI, and Symptoms ontology \cite{symp_ontology,schriml2022,schriml2012,chebi2006,chebi2016,chebi2008,geneontology2000}, as the ground truth to verify the term's identity. When a term is not found in an ontology, it is said to be ``unverified.'' (``Unverified'' does not mean ``invalid'' as the ontology may not be complete for some terms.) The second task is to verify the associations using the PubMed database as an authentic source of biomedical knowledge. 

\textbf{Task 2 (Association reliability verification):} This task is to verify associations between terms (i.e., named entities denoted by the terms), represented as a binary relation. Mathematically the association is expressed as \( R_{AB} \subseteq A \times B \), where \( A \) is a set of terms belonging to one category, such as diseases, and \( B \) is a set of terms from another category, such as genes, drugs, and symptoms. For instance, the association between diseases and genes can be expressed as \( R_{DG} \subseteq D \times G \), where \( D \) is the set of disease terms and \( G \) is the set of gene terms. A pair \( (d, g) \in R_{DG} \) signifies that a particular gene, \( g \), is associated with a specific disease, \( d \). This formal representation provides a structured way to model associations between different categories of terms.

\textbf{Task 3 (Association consistency verification): }
%{\color{red}{\textbf{Ahmed, do we have two tasks or three tasks in our work? If three tasks (I think so), can we also discuss the third task in this section, Section \ref{sec:methods}, and Section \ref{sec:discussion}? The third task seems to have ChatGPT generate simulated articles (i.e., PubMed abstracts) and test the ``consistency'' of the generated articles across different LLM models. This task is mentioned in the Highlights and the Results section (Section \ref{sec:results}).}}}
This task is to verify the consistency of knowledge (centered on associations between biomedical terms) among different ChatGPT models. It generates simulated abstracts from the different models and compares the associations detected from the simulated abstracts with the original associations generated in Task 2.

%%%%%%%%%%%%%%%%%%%%%%%%%%%%%%%%%%%%% Experimental Methods %%%%%%%%%%%%%%%%%%%%%%%%%%%%%%%%%%%%%
%                                     Experimental Methods                                      %
% -----------------------------------------------------------------------------------------------

\section{Experiment Methods}\label{sec:methods}

\subsection{Data generation via prompt engineering}
To test the knowledge generation capabilities of ChatGPT, we used means of prompt engineering via the APIs. The purpose was to instruct ChatGPT to generate various types of disease-centric term associations to enable the verification process. These term associations are the basic building blocks of more complex forms of knowledge represented in the knowledge networks. Generating and verifying various types of associations makes the task of knowledge verification easy and efficient by decomposing the verification tasks to fine-grained unit of term associations, thereby reducing the effort to build a large and complex knowledge graph. Specifically, we instructed ChatGPT to generate 5000 associations between disease on one side, and gene, symptom, and drug, respectively, on the other side. Recall that the main idea for verification is to verify (1) whether the terms of the associations are verifiable from the corresponding ontology and (2) whether the actual association instances are rooted in the literature. To this end, the prompt included generating pairs of verifiable ontology-terms with their IDs. 

For the purpose of smooth processing, the prompt also instructed ChatGPT to format the output in JSON format, which was then validated and saved to a file. Figure\,\ref{fig:association-samples} shows samples containing a few records from each of the three types of associations generated. Specifically, Figure\,\ref{fig:disease-drug} shows three diseases (breast cancer, asthma, and hypercholesterolemia) and the associated three drugs (Carbamazepine, Zidovudine, and Fluoxetine); Figure\,\ref{fig:disease-gene} and Figure\,\ref{fig:disease-symptom} show three genes (ACE, INS, and IL4) and three disease symptoms (headache, frequent urination, and shortness of breath), respectively.
\begin{figure}[H]
    \centering
    % Subfigure1: Disease-Drug Associations
    \begin{subfigure}[b]{0.65\textwidth}
        \centering
        \begin{tcolorbox}[colback=blue!5!white, colframe=blue!75!black, title=Disease-Drug Associations]
        \begin{itemize}
            \item \textbf{Disease:} \texttt{DOID:1612 (Breast Cancer)}\\
                  \textbf{Drug:} \texttt{ChEBI:33897 (Carbamazepine)}
            \item \textbf{Disease:} \texttt{DOID:2841 (Asthma)}\\
                  \textbf{Drug:} \texttt{ChEBI:10177 (Zidovudine)}
            \item \textbf{Disease:} \texttt{DOID:0050425 (Hypercholesterolemia)}\\
                  \textbf{Drug:} \texttt{ChEBI:5118 (Fluoxetine)}
        \end{itemize}
        \end{tcolorbox}
        \caption{Disease-drug associations.}
        \label{fig:disease-drug}
    \end{subfigure}
    \smallskip\\
    % Subfigure2: Disease-Gene Associations
    \begin{subfigure}[b]{0.65\textwidth}
        \centering
        \begin{tcolorbox}[colback=green!5!white, colframe=green!75!black, title=Disease-Gene Associations]
        \begin{itemize}
            \item \textbf{Disease:} \texttt{DOID:9351 (Hypertension)}\\
                  \textbf{GO term:} \texttt{GO:0002027 (Blood Pressure Regulation)}\\
                  \textbf{Gene/Protein:} \texttt{ACE (Angiotensin-Converting Enzyme)}
            \item \textbf{Disease:} \texttt{DOID:11723 (Diabetes Mellitus Type 2)}\\
                  \textbf{GO term:} \texttt{GO:0006006 (Glucose Metabolic Process)}\\
                  \textbf{Gene/Protein:} \texttt{INS (Insulin)}
            \item \textbf{Disease:} \texttt{DOID:2841 (Asthma)}\\
                  \textbf{GO Term:} \texttt{GO:0050707 (Regulation of Cytokine Secretion)}\\
                  \textbf{Gene/Protein:} \texttt{IL4 (Interleukin-4)}
        \end{itemize}
        \end{tcolorbox}
        \caption{Disease-gene associations.}
        \label{fig:disease-gene}
    \end{subfigure}
    \smallskip\\
    % Subfigure3: Disease-Symptom Associations
    \begin{subfigure}[b]{0.65\textwidth}
        \centering
        \begin{tcolorbox}[colback=red!5!white, colframe=red!75!black, title=Disease-Symptom Associations]
        \begin{itemize}
            \item \textbf{Disease:} \texttt{DOID:9352 (Essential Hypertension)}\\
                  \textbf{Symptom:} \texttt{SYMPTOM:1001 (Headache)}
            \item \textbf{Disease:} \texttt{DOID:11723 (Diabetes Mellitus)}\\
                  \textbf{Symptom:} \texttt{SYMPTOM:1002 (Frequent Urination)}
            \item \textbf{Disease:} \texttt{DOID:2841 (Asthma)}\\
                  \textbf{Symptom:} \texttt{SYMPTOM:1003 (Shortness of Breath)}
        \end{itemize}
        \end{tcolorbox}
        \caption{Disease-symptom associations.}
        \label{fig:disease-symptom}
    \end{subfigure}

    \caption{Overview of disease associations with drugs, genes, and symptoms.}
    \label{fig:association-samples}
\end{figure}

Algorithm\,\ref{alg:prompt_eng_assoc} shows a one-shot prompt engineering for the task of generating disease-symptom associations. 

\begin{algorithm}
\caption{One-shot prompt engineering for generating disease-symptom associations record in JSON format.}\label{alg:prompt_eng_assoc}
\begin{small}
\begin{algorithmic}[1]

\REQUIRE 
    \texttt{Model M, shot s\_1, Number of associations N} \\
    \texttt{s\_1} $\gets$ \{\\
    \hspace*{1.5em} \texttt{"DOID:11734"}: \texttt{"Epistaxis"},\\
    \hspace*{1.5em} \texttt{"SYMPTOM:1080"}: \texttt{"Nosebleed"}\}
\ENSURE
    \texttt{Valid and structured JSON output R containing N DOID-SYMPTOM associations.}

\STATE \textbf{\textsf{DEFINE}} \texttt{prompt P:} \\
\STATE \hspace{1em} \texttt{P $\gets$ ``You are an assistant that generates 10 DOID-SYMPTOM term associations in a structured JSON format. Ensure the JSON is valid and correctly formatted for parsing. Provide one example in the following format:''} \\
\STATE \hspace{2em} \texttt{P $\gets$ P + s\_1}

\STATE \textbf{\textsf{INITIALIZE}} \texttt{response request to model M:}
\STATE \hspace{1em} \texttt{R $\gets$ M.generateResponse(P, model = "gpt-4o")}

\STATE \textbf{\textsf{PROCESS}} \texttt{response R:}
\IF{\texttt{R} is a valid JSON format}
    \STATE \textbf{Output} \texttt{R}
\ELSE
    \STATE \textbf{Report error}: \texttt{``Invalid JSON format in response.''}
\ENDIF
\RETURN \texttt{R}
\end{algorithmic}
\end{small}
\end{algorithm}

% \begin{multicols}{2}
% % First Box
% \begin{tcolorbox}[colback=blue!5!white, colframe=blue!75!black, title=Disease-Drug Associations]
% \label{box:disease-drug}
% \begin{itemize}
%     \item \textbf{DOID:1612} Breast Cancer\\
%           \textbf{ChEBI:33897} Carbamazepine
%     \item \textbf{DOID:2841} Asthma\\
%           \textbf{ChEBI:10177} Zidovudine
%     \item \textbf{DOID:0050425} Hypercholesterolemia\\
%           \textbf{ChEBI:5118} Fluoxetine
%     \item \textbf{DOID:9256} Alzheimer's\\
%           \textbf{ChEBI:39548} Atorvastatin)          

% \end{itemize}
% \end{tcolorbox}

% % Third Box
% \begin{tcolorbox}[colback=orange!5!white, colframe=orange!75!black, title=Disease-Symptom Associations]
% \label{box:disease-symp}
% \begin{itemize}
%     \item \textbf{DOID:9352:} Essential Hypertension\\
%           \textbf{SYMPTOM:1001:} Headache
%     \item \textbf{DOID:11723:} Diabetes Mellitus\\
%           \textbf{SYMPTOM:1002:} Frequent Urination
%     \item \textbf{DOID:2841:} Asthma\\
%           \textbf{SYMPTOM:1003:} Shortness of Breath
% \end{itemize}
% \end{tcolorbox}

% \end{multicols}

\begin{table}[htbp]
\centering
\caption{Datasets generated and used in the experiments. It shows the total number of associations for each of three ontology pairings DOID-SYMP, DOID-GO, and DOID-CHEBI. These datasets were validated against trusted ontologies to ensure precision and accuracy. They are designed to support various bioinformatics applications, including disease annotation, genetic process mapping, and drug-disease linkage studies.}
\label{tab:gpt_generated_associatons}
\begin{small}
\begin{tabular}{@{}lc@{}}
\toprule
\textbf{Dataset} & \textbf{Total \#} \\ \midrule
ChatGPT DOID-SYMP JSON Pairs   & 5466               \\
ChatGPT DOID-GO JSON Pairs     & 5008               \\
ChatGPT DOID-CHEBI JSON Pairs  & 2625               \\ \bottomrule
\end{tabular}
\end{small}
\end{table}

\subsection{Term correctness verification against biomedical ontologies}
In this work, we address four types of terms that make up the three types of associations that can be represented formally as binary relations.

\begin{itemize}
\item[1.] \textbf{Disease-drug association:} This type of association is represented as \( R_{DD} \subseteq D \times Dr \), where \( Dr \) denotes the set of drugs. It signifies the relationship between diseases and the drugs used for their treatment or management.

\item[2. ] \textbf{Disease-symptom association:} This association is denoted as \( R_{DS} \subseteq D \times S \), where \( S \) represents the set of symptoms. It captures the relationship between diseases and their corresponding symptoms.

\item[3.] \textbf{Gene-process association:} This is expressed as \( R_{GP} \subseteq G \times P \), where \( P \) represents the set of genetic processes. It reflects the relationship between specific genes and the biological processes they are involved in.
\end{itemize}

\begin{figure}[H]
    \centering
    \includegraphics[scale=0.75]{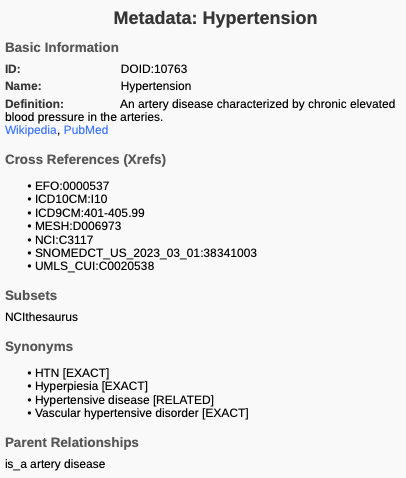}
    \caption{The term ``Hypertension'', identified as DOID:10763 in the DOID ontology, includes a metadata item for synonyms. %A term-matching
    The term verification algorithm (Algorithm\,\ref{alg:term_verification}) utilizes this synonyms field to semantically verify the legitimacy of terms generated by ChatGPT.}
    \label{fig:doid-synonym}
\end{figure}

The terms generated by the model are verified against a specialized ontology. For the diseases, the terms are verified against the the Human Disease Ontology (DOID), the drugs are verified using the Chemical Entities of Biological Interest (ChEBI) ontology, the genetic knowledge is  verified using the Gene Ontology (GO), whereas the symptoms are verified using the The Symptom Ontology (SYMP). Since the various ontologies maintain term names and synonyms, the term verification is semantic in nature. Table\,\ref{tab:gpt_generated_associatons} describes three datasets of associations among disease-symptom, disease-gene, and disease-drug, respectively. Figure\,\ref{fig:doid-synonym} shows a proof of a synonym metadata item that describe a DOID:10763 ontology term \cite{disease_ontology_doid10763} known as hypertension and various synonyms namely: (HTN, Hyperpiesia, Hypertensive disease, Vascular hypertensive disorder).

Algorithm\,\ref{alg:term_verification} shows the steps for term verification from the domain ontology. Each ontology term may have a list of synonyms whenever applicable. The algorithm needs to search both the list of terms and the list of synonyms in the ontology to ensure sound and fair verification.

\begin{algorithm}[htbp]
\caption{Term verification Using domain-specific ontology.}
\label{alg:term_verification}
\begin{small}
\begin{algorithmic}[1]
\REQUIRE Term instance \( t \), Term type \( \tau \), Domain ontology \( \mathcal{O} \)
\ENSURE \texttt{VERIFIED} or \texttt{UNVERIFIED}
\STATE Let \( L_t \) be the list of terms in the ontology \( \mathcal{O} \)
\FORALL{terms \( t_i \in L_t \)}
    \IF{\( t = t_i \)}
        \RETURN \texttt{VERIFIED}
    \ELSE
        \STATE Let \( L_s \) be the list of synonyms for \( t \)
        \FORALL{synonyms \( s_j \in L_s \)}
            \IF{\( t = s_j \)}
                \RETURN \texttt{VERIFIED}
            \ENDIF
        \ENDFOR
    \ENDIF
    \RETURN \texttt{UNVERIFIED}
\ENDFOR
\end{algorithmic}
\end{small}
\end{algorithm}

% ---------------------------------------------------------------------------------------------------

\subsection{Association reliability verification against biomedical literature}

Since the specialized ontologies are term-based resources, they can only support the verification of the terms that make up the associations. Therefore, verification of associations goes beyond the ontology. It is a common practice for scientists to verify biomedical knowledge from the scientific literature. This requires searching digital repositories such as PubMed \cite{pubmed} and manually verifying whether an evidence holds. The rational behind using PubMed for verification is that PubMed has been recently improved to respond to certain information needs, specifically related to evidence-based medicine and association discovery \cite{jin2024pubmed}.

The manual process, though may produce highly accurate results, is labor-intensive and slow in nature. With the large amount of content generated by ChatGPT and other GenAI tools, this process is infeasible. %obsolete. 
Here, we present an algorithmic approach that offers an automatic means to verify the generated associations. The algorithm is designed to search contextual dataset of biomedical abstracts for a certain association type. To gather this dataset, we performed a contextual search for the three associations: (1) disease and drug, (2) disease and gene, and (3) disease and symptom. Each dataset was used as the basis to verify the corresponding association. %That is, the disease-drug dataset was used to verify the disease-drug associations. 

The steps of association verification are outlined in Algorithm\,\ref{alg:association_verification}. The notion of association verification here can be defined as a co-occurrence of the association terms in one of more PubMed abstracts. The number of PubMed abstracts that contain an association is called  ``coverage''. The number of associations supported by coverage offers a score for a given type of association using the underlying dataset. 

\begin{algorithm}[H]
\caption{Association verification using literature.}
\label{alg:association_verification}
\begin{small}
\begin{algorithmic}[1]
\REQUIRE List of association term pairs \( L_p \); Dataset of literature \( D \)
\ENSURE List of verified associations and their hit ratios, \( L_v \)
\STATE Initialize \( L_v \leftarrow \emptyset \)
\FORALL{association term pairs \( P \equiv (p_i, p_j) \in L_p \)}
    \STATE Initialize \( \text{counter} \leftarrow 0 \)
    \FORALL{abstracts \( B \in D \)}
        \IF{\( p_i \in B \) \textbf{and} \( p_j \in B \)}
            \STATE \( \text{counter} \leftarrow \text{counter} + 1 \)
        \ENDIF
    \ENDFOR
    \STATE compute hit\_ratio \( h_P \leftarrow \frac{\text{counter}}{\lvert L_P \rvert} \)
    \IF {\( h_P > 0 \)}
        \STATE append \( (P, h_P) \) to \( L_v \)
    \ENDIF
\ENDFOR
\RETURN \( L_v \)
\end{algorithmic}
\end{small}
\end{algorithm}

% \subsection{Association consistency verification in generated biomedical abstracts across different models}

\section{Experiment Results}\label{sec:results}
We present the results of the experiments in evaluating ChatGPT's capabilities in the following key tasks: 
\begin{itemize}
    \item Verifying the correctness of the biomedical terms that make up the associations (i.e., disease, symptom, drug, and genes);
    \item Verifying the associations linkage against biomedical literature from different periods;
    \item Testing the randomness of ChatGPT by generating simulated articles using various ChatGPT models.
\end{itemize}

\subsection{Task 1 -- Verification of the correctness of biomedical terms}
We evaluated the names of the three types of associations generated disease-drug, disease-symptom, and disease-gene/genetic process that made up the ChatGPT-generated associations using domain-specific ontologies as ground truth. The verification of the terms that make up the generated associations was checking against the DOID ontology for disease terms, the ChEBI ontology for the drug terms, the SYMPTOM ontology for the symptom terms,  and the GO ontology for the genetic terms (gene names and genetic processes). The encoding of those ontology offer means of literal and semantic matching, which offers fair means of comparisons. For instance, the (``hypertension'') disease term in the DOID ontology  (``DOID:10763'') \cite{disease_ontology_doid10763}. Additionally, the ontology entity of this term also includes the list of synonyms (``HTN [EXACT], hyperpiesia [EXACT], hypertensive disease [RELATED], vascular hypertensive disorder [EXACT]''), which are also checked during the algorithmic process. Hence, the claim of a semantic verification process. 

\subsection{Tasks 1.1 -- Verification of disease terms}
The task of of generating disease terms was common across three types of associations. The verification result of the terms in the three types of associations are as follows.
%\begin{enumerate}
\begin{itemize}
    \item%[1.]
    For disease-drug associations, the literal matching process verified 93\% of disease terms, while the semantic matching verified 87\% of the generated names. Combined, 98\% of disease names were successfully verified.  
    \item%[2.]
    For disease-symptom associations, the literal matching verified 97\% of the disease terms, while the semantic matching verified 82\% of the generated terms. Combined, 99\% of disease names were successfully verified.  
    \item%[3.] 
    For disease-gene associations, the literal matching verified 88\% of the disease terms, while semantic matching verified 97\% of the generated terms. Due to the high percentage of verification, the task of a combined matching was omitted. %from literal and semantic was ignored.
\end{itemize}
%\end{enumerate}

\subsubsection{Tasks 1.2 -- Verification of non-disease terms}
Here we summarize the results of ChatGPT generating correct drugs, symptoms, genes, and genetic processes as part of the associations:
\begin{itemize}
    \item%[1.]
    \textbf{Drug names:} the literal matching verified 90\% of drug names, with an 90\% verified through synonym matching. The combined verification rate was 91\%.  
    \item%[2.]
    \textbf{Symptom names:} literal matching verified 49\% of symptom names, with an additional 25\% verified through semantic matching. The combined verification rate was 61\%.  
    \item%[3.]
    \textbf{Genetic processes and gene names:} the verification resulted in the verification 80\% of the gene names and 97\% of the genetic processes.
\end{itemize}
%\textbf{**I would keep a consistent writing style for all cases instead of using different words (for the same meaning) for different cases.**}

\begin{figure}[H]
    % \centering
    \includegraphics[scale=.65]{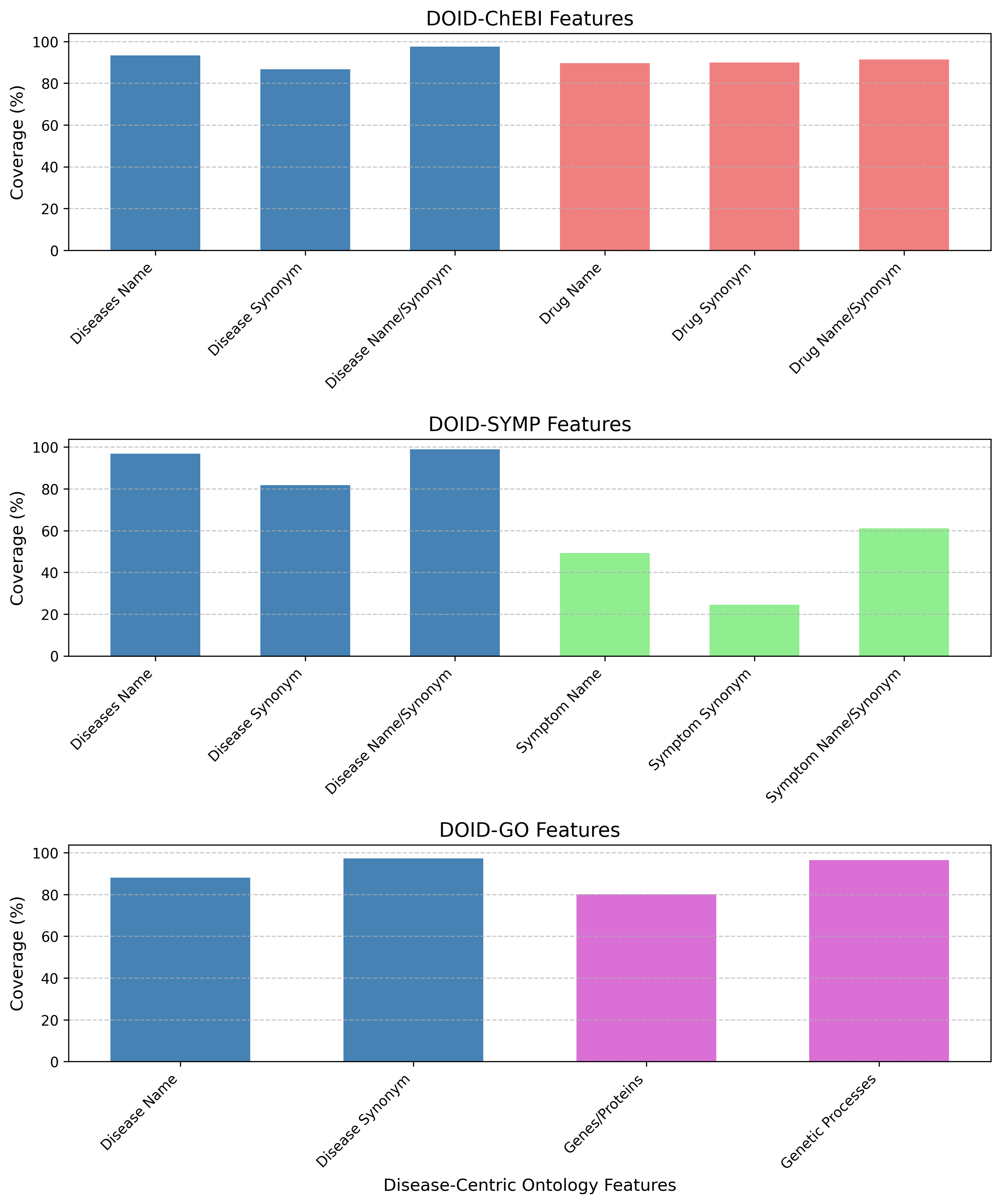}
    \caption{Accuracy of DOID-ChEBI, DOID-SYMP, and DOID-GO associations across various features.}
    \label{fig:doid_associations_task1}
\end{figure}

These findings demonstrate the strong capability of ChatGPT in generating biomedical terms that align closely with biomedical ontology as one of the most authentic sources of ground truth. Even in the case of the symptom terms, where performance was notably lower, this does %mean
pose a significant concern from the point of view of our study. In Section \ref{sec:task2}, the results will show that, although not identified in the specialized ontology, there was a noticeable improvement when searching the literature for association links which also included symptom terms. Table\,\ref{tab:task1_stats} captures the statistics that summarize this task, and the results are shown in Figure\,\ref{fig:doid_associations_task1}.

\begin{table}[htbp]
\centering
\caption{Verification of entity names and types of ChatGPT-generated associations of disease-symptoms-drug-gene using biomedical ontologies (DOID for diseases, SYMP for disease symptom names, ChEBI for drug names, and GO for genetic processes and gene names). The disease-centric links generated were DOID--ChEBI, DOID--SYMP, and DOID-GO.}
\label{tab:task1_stats}
\renewcommand{\arraystretch}{1.2}
\begin{small}
\begin{tabular}{l|l|c}
\toprule
\textbf{Category}                  & \textbf{Feature Verified}             & \textbf{Accuracy (\%)} \\ \hline
\multirow{3}{*}{DOID-ChEBI associations} 
                                   & Diseases name               & 93.37          \\ 
                                   & Disease synonym             & 86.70          \\ 
                                   & Disease name/synonym        & 97.60          \\ 
                                   & Drug name                   & 89.52          \\ 
                                   & Drug synonym                & 89.98          \\ 
                                   & Drug name/synonym           & 91.43          \\ \midrule
\multirow{6}{*}{DOID-SYMP associations} 
                                   & Diseases name               & 96.83          \\ 
                                   & Disease synonym             & 81.87          \\ 
                                   & Disease name/synonym        & 98.87          \\ 
                                   & Symptom name                & 49.29          \\ 
                                   & Symptom synonym             & 24.50          \\ 
                                   & Symptom name/synonym        & 61.14          \\ \midrule
\multirow{5}{*}{DOID-GO associations} 
                                   & Disease name                & 88.12          \\ 
                                   & Disease synonym             & 97.36          \\ 
                                   & Genes/proteins              & 80.21          \\ 
                                   & Genetic processes           & 96.47          \\ \bottomrule
                                   % & Same Entity Processes-Gene  & 98.74          \\ \bottomrule
\end{tabular}
\end{small}
\end{table}

%%%%%%----------------------------------------------------------------------------------------%%%%%%%%

\subsection{Task 2: Verification of the reliability of biomedical association links}\label{sec:task2}
While domain-specific ontologies offer verification of individual entities of the associations, they do not offer the means of verifying the actual relationships between the individual terms. To address this limitation, we utilized the biomedical literature as another reputablesource of ground truth to verify the various associations. Thus, we constructed three datasets, each to verify one of the three types of associations, that is, disease-drug, disease-genetic information, and disease-symptom. Because the GO ontology does not reference the direction of the gene terms in the ontology, %but
in another resource known as the GO ontology annotation \cite{camon2004gene,yon2008use,gene2012gene}, we prompted ChatGPT to produce associations that link a disease to two pieces of genetic  knowledge --- the GO term, which is the genetic process itself, and the gene name, which is the product of the actual process. This requires the validation of mapping the biological process to its product, which is also included in the results.  

In addition, to examine the effect of publication longevity, each data set was divided into three 5-year periods spanning 2009 to 2024. The verification processes consider the literature coverage as means of being viable. The premise is that if an association occurs frequently in a set of PubMed abstracts then this makes it ``verified''. The research stops at this step and does not further explore the semantics of such associations. As mentioned earlier, if an association was not covered in the literature then this means it is yet to be verified. In other words, it does not label an association as correct or incorrect. The experiments revealed a consistent trend of improved verification coverage across all types of association over the three periods. Specifically:
\begin{enumerate}
    \item \textbf{Disease-drug association} is verified with coverage rates of 86\%, 88\%, and 90\%, respectively, over the three periods.
    \item \textbf{Disease-gene association} is verified with coverage rates of 83\%, 83\%, and 89\%, respectively, over the three periods.
    \item \textbf{Disease-symptom association} is verified with coverage rates of 49\%, 53\%, and 62\%, respectively, over the three periods.
    \item \textbf{Genetic process-gene association} is verified with the coverage improved from 23\% in the first period to 83\% and 89\% in the second and third period, respectively, over the three periods.
\end{enumerate}

Additionally, we further analyzed the frequency of publications supporting each type of association to measure the level of support from the literature over time. The results summarized in Table\,\ref{tab:task2_stats} indicate an increasing trend of publication support over the three periods of time (2009-2014, 2015-2019, and 2020-2024):
\begin{itemize}
    \item \textbf{Disease-gene association} is supported by an average of 90, 82, and 214 publications, respectively, over the three periods.
    \item \textbf{Disease-drug association} is supported by an average of 31, 56, and 107 publications, respectively, over the three periods.
    \item \textbf{Disease-symptom association} exhibits the lowest support, with an average of 9, 14, and 31 publications, respectively, over the three periods. However, it still follows an upward trend.
\end{itemize}

\begin{table}[htbp]
    \centering
    \caption{Average co-occurrences across time periods for disease associations.}\label{tab:task2_stats}
    \begin{small}
    \begin{tabular}{@{}llc@{}}
        \toprule
        \textbf{Association Type}      & \textbf{Time Period} & \textbf{Average Frequency} \\ 
        \midrule
        Disease -- gene           & 2009--2014           & 90.48                          \\
                                       & 2015--2019           & 82.54                          \\
                                       & 2020--2024           & 214.38                         \\
        \midrule
        Disease -- drug           & 2009--2014           & 30.71                          \\
                                       & 2015--2019           & 56.37                          \\
                                       & 2020--2024           & 107.34                         \\
        \midrule
        Disease -- symptom        & 2009--2014           & 9.17                           \\
                                       & 2015--2019           & 13.96                          \\
                                       & 2020--2024           & 31.40                          \\
        \bottomrule
    \end{tabular}
    \end{small}
\end{table}

\begin{figure}[H]
    \centering
    \includegraphics[scale=.45]{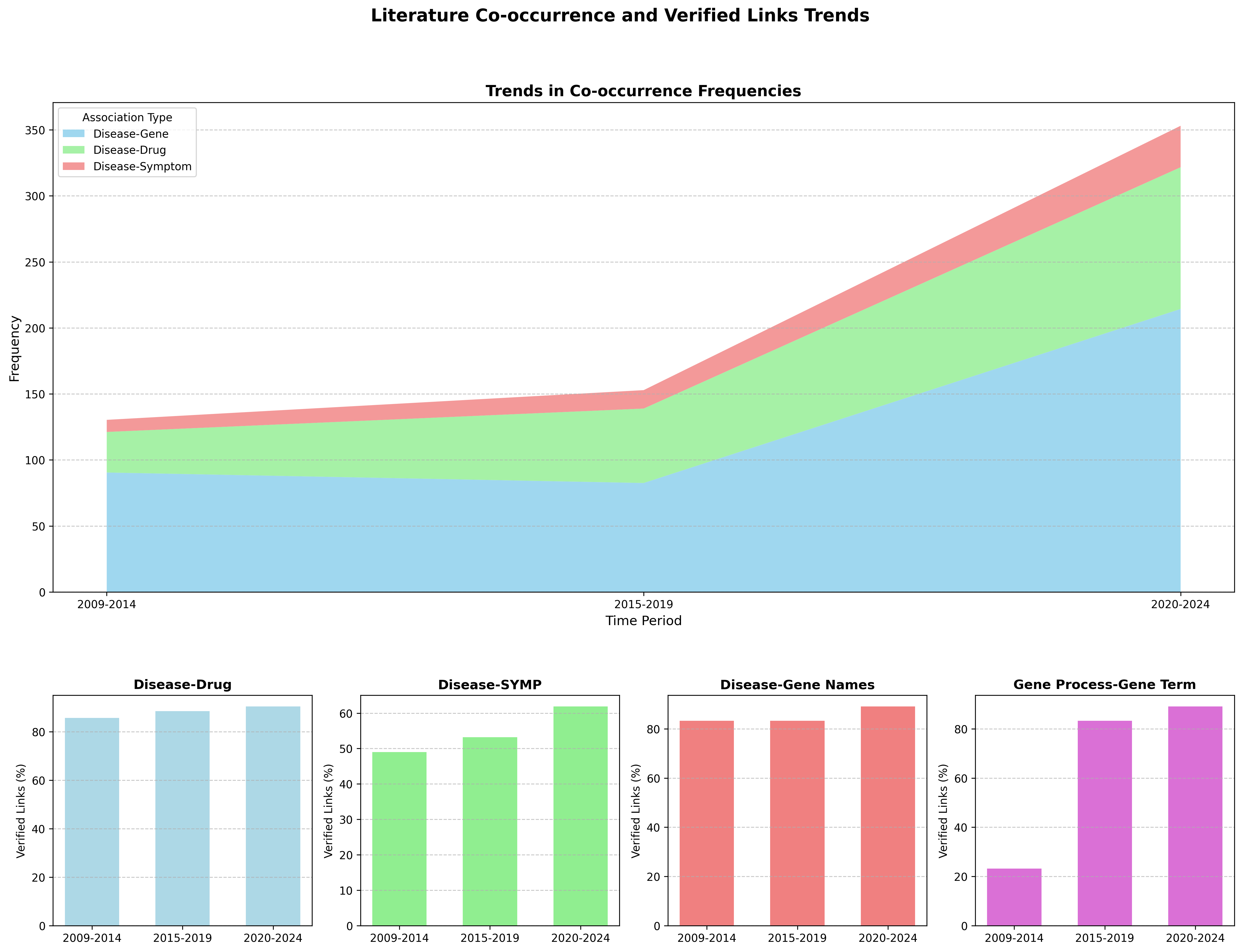}    
    \caption{Combined visualization of average co-occurrences (line chart) and literature co-occurrence statistics (bar charts). The top plot shows trends in co-occurrences over time, while the bottom plot compares verified and unverified links for different association types.}
    \label{fig:task2_task3_combined}
\end{figure}

The results from the two tasks demonstrate the high accuracy of the biomedical terms that make up the ChatGPT-generated associations, and also show an increasing coverage trend over time as captured in Table~\ref{tab:task2.1_stats}. The trend suggests that recent publications contribute more robustly to the verification of biomedical associations as shown in Figure\,\ref{fig:task2_task3_combined}. On the one hand, the ChatGPT associations depict an evolutionary picture of the knowledge accumulated in biomedical literature over time. On the other hand, the verification process may also reflect on the type of knowledge ChatGPT may contribute and an indication of how the pre-training processes of ChatGPT has taken place. 

\begin{table}[htbp]
    \centering
    \caption{Literature co-occurrence statistics for disease-drug, disease-symptom, disease-gene name, and gene process associations.}\label{tab:task2.1_stats}
    \begin{small}
    \begin{tabular}{@{}llcc@{}}
        \toprule
        \textbf{Association Type}      & \textbf{Time Period} & \textbf{Unverified Links (\%)} & \textbf{Verified Links (\%)} \\ 
        \midrule
        Disease -- drug                 & 2009--2014           & 14.29                         & 85.71                        \\
                                       & 2015--2019           & 11.53                         & 88.47                        \\
                                       & 2020--2024           & 9.52                          & 90.48                        \\
        \midrule
        Disease -- symptom                 & 2009--2014           & 51.02                         & 48.98                        \\
                                       & 2015--2019           & 46.83                         & 53.17                        \\
                                       & 2020--2024           & 38.08                         & 61.92                        \\
        \midrule
        Disease -- gene           & 2009--2014           & 16.74                         & 83.26                        \\
                                       & 2015--2019           & 16.74                         & 83.26                        \\
                                       & 2020--2024           & 10.85                         & 89.15                        \\
        \midrule
        Gene process -- gene term       & 2009--2014           & 76.84                         & 23.16                        \\
                                       & 2015--2019           & 16.74                         & 83.26                        \\
                                       & 2020--2024           & 10.85                         & 89.15                        \\
        \bottomrule
    \end{tabular}
    \end{small}
\end{table}

%%%%%%------------------------------------------------------------------------------------%%%%%%%%

\subsection{Task 3: Verification of the association consistency against ChatGPT-simulated Abstracts by various models}
\begin{figure}[H]
    \centering
    \includegraphics[scale=.45]{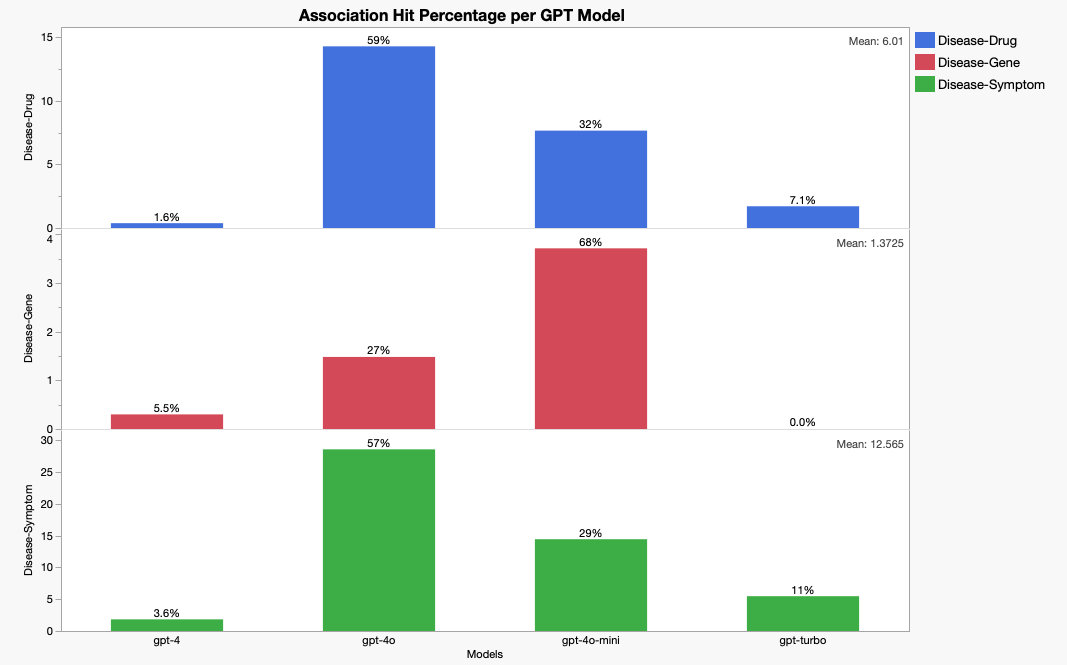}    
    \caption{A model performance comparison, as means of self-consistency,  for the associations generated from independent prompts}
    \label{fig:task3-model-comparison}
\end{figure}

%To assess the degree of randomness of ChatGPT generating the association studied in this work,
To assess how consistent (or random) the ChatGPT-generated associations are, we generated disease-centric simulated abstracts using various ChatGPT models. Specifically, we prompted four ChatGPT models --- ChatGPT-4, ChatGPT-4turbo, ChatGPT-4o, and ChatGPT-4omini --- to generate simulated abstracts centered on human diseases. Due to the computational cost of this task, we limited the generation to approximately 5,000 abstracts per model. Figure~\ref{fig:task3-model-comparison} shows number of hits per model for three types os associations in three layers: the top layer in blue is for disease-drug, the middle layer in red shows the drug-genes, while the bottom layer in green is for the disease-symptom.   The verification process is summarized in Table\,\ref{tab:task3_model_comparison}. 
\begin{itemize}
    \item \textbf{Disease-drug association} achieved coverage rate of (1\% to 15\%);
    \item \textbf{Disease-gene association} achieved coverage rate of (1\% to 4\%); 
    \item \textbf{Disease-symptom association} achieved coverage rate of (2\% to 29\%). 
\end{itemize}

\begin{table}[htbp]
\centering
\caption{Combined statistics of disease-drug, disease-gene, and disease-symptom associations checked against the ChatGPT-generated simulated biomedical abstracts.}
\label{tab:task3_model_comparison}
\begin{small}
\begin{tabular}{@{}lccc@{}}
\toprule
\textbf{Model}                  & \textbf{Count} & \textbf{Percentage (\%)} & \textbf{Association Type} \\ \midrule
chatgpt\_4\_model\_count        & 10             & 0.38                     & Disease-Drug              \\ 
model\_turbo\_chatgpt\_count    & 45             & 1.71                     & Disease-Drug              \\ 
chatgpt-4o-mini-model           & 201            & 7.66                     & Disease-Drug              \\ 
chatgpt-4o-model                & 375            & 14.29                    & Disease-Drug              \\ \midrule
chatgpt\_4\_model\_count        & 15             & 0.30                     & Disease-Gene              \\ 
model\_turbo\_chatgpt\_count    & 0              & 0.00                     & Disease-Gene              \\ 
chatgpt-4o-mini-model           & 186            & 3.71                     & Disease-Gene              \\ 
chatgpt-4o-model                & 74             & 1.48                     & Disease-Gene              \\ \midrule
chatgpt\_4\_model\_count        & 100            & 1.83                     & Disease-Symptom           \\ 
model\_turbo\_chatgpt\_count    & 299            & 5.47                     & Disease-Symptom           \\ 
chatgpt-4o-mini-model           & 788            & 14.42                    & Disease-Symptom           \\ 
chatgpt-4o-model                & 1560           & 28.54                    & Disease-Symptom           \\ \bottomrule
\end{tabular}
\end{small}
\end{table}

\begin{figure}[H]
    \centering
    \includegraphics[scale=.4]{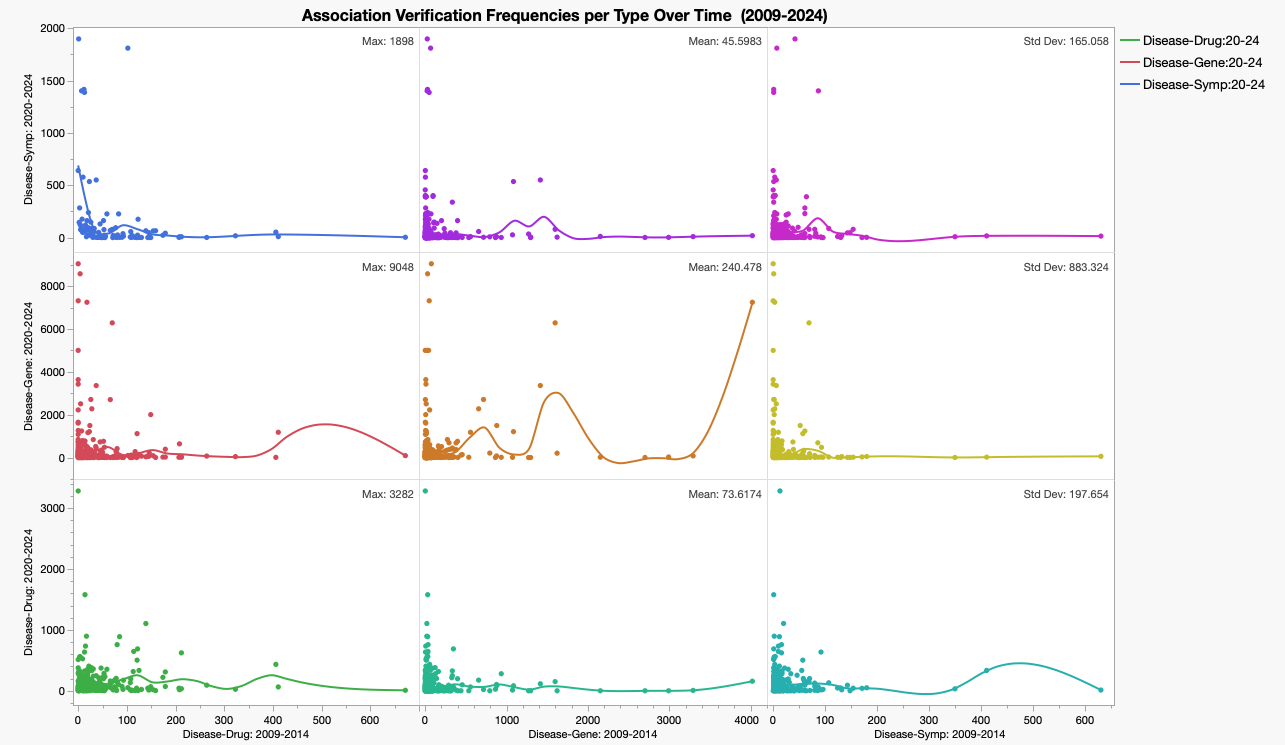}    
    \caption{Time and summary statistics of the association coverage for three types(disease-symptom), (disease-drug), and (disease-gene).}
    \label{fig:sum_stat}
\end{figure}

While these coverage rates appear modest compared to benchmarks against much larger biomedical literature datasets (spanning 250,000 to 650,000 abstracts), the disparity is likely attributableto the smaller dataset size in this evaluation. Notably, the disease-symptom associations exhibited the highest match rates, a result that contrasts with their lower performance in comparisons against biomedical literature datasets. These findings underscore the potential of ChatGPT models to identify consistent disease-symptom links, though further investigation is required to validate the correctness of these associations in more focused contexts. Figure\,\ref{fig:sum_stat} shows the evaluation of associations across the various models and summary statistics for each type over three 5-year publication periods. 

%\section{Observations and Anecdotal Evidence}
\section{Discussion} \label{sec:discussion}

In this work, we tested the capabilities of ChatGPT to generate biomedical associations as building blocks for more complex data models such as biomedical networks and knowledge graphs. Specifically, we designed a prompt-engineering algorithm that produces human disease-centric associations in the context of symptoms, drugs, genetics. The algorithm prompted for generating association terms that match the corresponding specialized ontology, namely, DOID, ChEBI, SYMPTOM, and GO ontology. The prompt also provided a shot as an example of what is to be produced for a valid association. Each association was to be between two terms, a source and a target, where each term is encoded by a term ID and a name (or an instance of its synonyms). %As discussed in the methods (Section\,\ref{sec:methods}),% and the results, 
Note the research was driven by the verification of the terms that make up the generated associations using the mentioned ontologies. The associations were verified against the biomedical literature using instances of PubMed abstracts of different periods. Here, we discuss our observations along with some anecdotal evidence in each of the verification tasks performed.

\subsection{Term correctness verification}
The most striking observation is in the outcome of low coverage or symptom term verifications using SYMP ontology. Our manual analysis revealed several critical reasons for the challenges encountered when matching symptom terms from the ontology during the verification process: (1) the language discrepancy posed significant challenges. ChatGPT predominantly generates symptoms described in social or layman's terms that resonate more with general audiences rather than the specialist terminology employed in biomedical ontologies. Unlike drug terminology, which typically has multiple recognized names (generic, brand, etc.), symptoms frequently lack such standardized variations. Examples illustrating this challenge include terms like ``itchy blisters'' versus ``blister'', "seizures" instead of "febrile convulsion," "facial redness" instead of "inflammation," and "sad mood" versus ``depression''. (2) ChatGPT tended to generate overly verbose or too detailed symptom descriptions, complicating direct matching with succinct ontology terms or their synonyms. Examples include phrases such as "rapid weight loss" versus simply "weight loss," "swelling in joints," and "swollen lymph nodes," where simpler ontology terms exist. (3) instances occurred where multiple symptoms were combined into one description, preventing direct term-to-term matching. A notable example was combining ``bulls-eye/bull's eye'' (typically classified as a lesion) with "rash," thus complicating the ontology-based verification. (4) inconsistencies in punctuation and lack of synonym availability within the ontology further hindered symptom identification, underscoring the importance of advanced semantic matching mechanisms. These issues also contributed to the verification of disease-symptom associations. 

\subsection{Association reliability and consistecny verification} 
Association verification against the biomedical literature results in a positive outcome of very high coverage for some associations. This observation offers some confidence knowing that ChatGPT has certain knowledge that may be considered the fundamentals of science, common knowledge, or frequently studied associations; examples of such associations are the disease-drug association between diabetes mellitus and insulin, which was covered by 3000+ co-occurrences; the disease-gene association between breast cancer and ataxia telangiectasia mutated (ATM), where 9000 co-occurrences pointed to it that the ATM gene may cause the breast cancer. Generating such associations may summarize the basic building blocks in the human diseases, which is the ultimate objective of this study. It may also trigger incremental generation of associations while performing verification, to construct a more comprehensive human disease landscape.

\subsection{Association consistency verification}
When performing the association consistency verification we observed that some of the frequencies is only a single occurrence in a publication abstract. These low coverage associations may be coincidental in most cases. However, after investigating various instances we found, for example, that the association between non-small cell lung cancer and lapatinib as a disease-drug type, is proven correct. Further, we also found that the same drug was approved by the FDA to repurposed for breast cancer in combination with a chemotherapy drug known as capecitabine \cite{drugbank_DB01259},  which presents a piece of drug repurposing evidence. 

Other similar observations, the association between hypertensive heart disease and losartan also had a very weak coverage in the the literature. Losartarn as a standalone drug according to Xu et al.\cite{Xu_2009} or in combination with hydrochlorothiazide according to Suzuki et al.\cite{Suzuki_2015}, plays a significant role to treat cardiovascular diseases, including hypertension and heart diseases. Both publications became available in the period from 2009 to 2015. We hypothesize that ChatGPT may have generated this knowledge from another source of evidence was thought significant. It is important to study the associations of weak coverage and further explore such evidence to understand its value. 

When we tested if the publication date as a factor in the training of ChatGPT, we found that indeed the publication date played a role in the verification of the knowledge generated by ChatGPT. Specifically, we found more coverage for associations from publications published in recent years (2020-2024) than earlier. The coverage was significantly lower if the publication was old by a decade or more. However, the ChatGPT generated associations were only found in older literature than recent as in Xu et al., and Suzuki et al. \cite{Xu_2009,Suzuki_2015}. 

The way we used different ChatGPT models to generate simulated abstracts and to measure the coverage also proved that ChatGPT's ability to generate associations was based on a certain ground, and it was not random. Although the overlap of associations in the generated abstract was low, it was a good test of trust. It is important to acknowledge that the abstracts were generated in the most generic way based on its pre-trained knowledge about human diseases, symptoms, genetics, and drugs. More importantly, the abstracts were also generated in an entirely independent process and produced a different type of output (i.e., structured associations). 

\section{Future Directions}
To address the symptom identification issues, we plan to introduce an LLM-driven solution to measure the semantic similarities among various terms. Leveraging self-consistency verification methods within a single model or comparing outputs across multiple models may further enhance the outcomes of symptom verification and subsequent disease-symptom associations. Additionally, extending this research to social media phenomena related to adverse drug reactions (ADRs) and side-effects as sources of real-world data may prove valuable.

We have tested our verification methods and algorithms on various ChatGPT models as initial stable tools. However, we plan to push the boundaries using open-source models to eliminate cost limitations. Some of the open-source LLMs we plan to use include Microsoft Phi-3, Meta Llama, QWen2.5, and TiniZero \cite{microsoft2024phi3,tinyzero,meta2024llama,qwen2.5,qwen2,tinyzero}, among others. We also plan to enhance the completely unsupervised approach for generating associations through prompt engineering. This endeavor may be achieved by generating associations from biomedical literature with the help of LLMs. Incorporating retrieval augmented generation (RAG) provides the most relevant publications to the query, and the query results may also be used for in-context prompt engineering, employing one document at a time to ensure factuality and explainability \cite{khan2024reinforcement}. We believe such approaches will increase confidence in the knowledge generated and the verification outcomes.

Another potential direction is to further investigate the interesting associations identified during our analysis. For instance, the association between (``non-small cell lung cancer'') and (``lapatinib'') could be further explored using retrieval augmented generation (RAG) and prompt engineering. In this case, we would inject verified associations, along with their corresponding full PubMed abstracts, into a series of prompts to explore the surrounding facts comprehensively. For example, we may include the verified association as part of a prompt along with relevant PubMed abstracts, instructing the LLM to generate knowledge regarding drug-target interactions, drug side effects, drug alternatives, mechanisms of action, drug-drug interactions, drug combinations, drug repurposing evidence, and other relevant associations related to non-small cell lung cancer. The verification may follow the process outlined in Algorithm \ref{alg:association_verification} and be further explored by constructing literature-driven heterogeneous mirror networks that could potentially reveal missing links once verified.

\section{Declaration of generative AI and AI-assisted}
During the preparation of this work the author(s) used ChatGPT in order to generate associations and simulated articles to produce the datasets of this work. The authors also used ChatGPT to perform LaTeX formatting to enhance the presentation of the work. However, the writing was written entirely by the authors and no GenAI tool was involved in the writing process. 

\section{Acknowledgments}
The work presented in this paper has been partly supported from IBM Faculty Award by the IBM Corporation. Any opinions, findings, and conclusions or recommendations expressed in this paper are those of the authors and do not necessarily reflect the views of the IBM Corporation. The authors thank Professor Luis Rocha and the members of the CACSI Laboratory of Binghamton University for the valuable discussion.  The authors would like to thank Northeastern University Maimi MSIS students Han Shao, Srivarini Mandali, and Mengxia Qiu for valuable discussions.  

\bibliographystyle{elsarticle-num} 
\bibliography{cas-refs}

\begin{thebibliography}{10}
\expandafter\ifx\csname url\endcsname\relax
  \def\url#1{\texttt{#1}}\fi
\expandafter\ifx\csname urlprefix\endcsname\relax\def\urlprefix{URL }\fi
\expandafter\ifx\csname href\endcsname\relax
  \def\href#1#2{#2} \def\path#1{#1}\fi

\bibitem{chatgpt2025}
{OpenAI}, {ChatGPT: Conversational AI}, \url{https://chatgpt.com/}, accessed:
  2025-01-12 (2025).

\bibitem{hamed2024safeguarding}
A.~A. Hamed, M.~Zachara-Szymanska, X.~Wu, Safeguarding authenticity for
  mitigating the harms of generative ai: Issues, research agenda, and policies
  for detection, fact-checking, and ethical ai, IScience (2024).

\bibitem{hamed2024detection}
A.~A. Hamed, X.~Wu, Detection of chatgpt fake science with the xfakesci
  learning algorithm, Scientific Reports 14~(1) (2024) 16231.

\bibitem{ciampaglia2015computational}
G.~L. Ciampaglia, P.~Shiralkar, L.~M. Rocha, J.~Bollen, F.~Menczer,
  A.~Flammini, Computational fact checking from knowledge networks, PloS one
  10~(6) (2015) e0128193.

\bibitem{luengo2020performance}
M.~Luengo, D.~Garc{\'\i}a-Mar{\'\i}n, The performance of truth: politicians,
  fact-checking journalism, and the struggle to tackle covid-19 misinformation,
  American Journal of Cultural Sociology 8~(3) (2020) 405.

\bibitem{nyhan2020taking}
B.~Nyhan, E.~Porter, J.~Reifler, T.~J. Wood, Taking fact-checks literally but
  not seriously? the effects of journalistic fact-checking on factual beliefs
  and candidate favorability, Political behavior 42 (2020) 939--960.

\bibitem{rodriguez2021debunking}
C.~Rodr{\'\i}guez-P{\'e}rez, F.~J. Paniagua-Rojano, R.~Magall{\'o}n-Rosa,
  Debunking political disinformation through journalists’ perceptions: An
  analysis of colombia’s fact-checking news practices, Media and
  Communication 9~(1) (2021) 264--275.

\bibitem{zeng2021automated}
X.~Zeng, A.~S. Abumansour, A.~Zubiaga, Automated fact-checking: A survey,
  Language and Linguistics Compass 15~(10) (2021) e12438.

\bibitem{guo2022survey}
Z.~Guo, M.~Schlichtkrull, A.~Vlachos, A survey on automated fact-checking,
  Transactions of the Association for Computational Linguistics 10 (2022)
  178--206.

\bibitem{lazarski2021using}
E.~Lazarski, M.~Al-Khassaweneh, C.~Howard, Using nlp for fact checking: A
  survey, Designs 5~(3) (2021) 42.

\bibitem{oshikawa2018survey}
R.~Oshikawa, J.~Qian, W.~Y. Wang, A survey on natural language processing for
  fake news detection, arXiv preprint arXiv:1811.00770 (2018).

\bibitem{anusree2022factorfake}
V.~Anusree, K.~Aarsha~Das, P.~Arya, K.~Athira, S.~Shameem, Factorfake:
  Automatic fact checking using machine learning models, in: Machine Learning
  and Autonomous Systems: Proceedings of ICMLAS 2021, Springer, 2022, pp.
  179--191.

\bibitem{khalil2021detecting}
A.~Khalil, M.~Jarrah, M.~Aldwairi, Y.~Jararweh, Detecting arabic fake news
  using machine learning, in: 2021 second international conference on
  intelligent data science technologies and applications (IDSTA), IEEE, 2021,
  pp. 171--177.

\bibitem{zhou2019physiological}
J.~Zhou, H.~Hu, Z.~Li, K.~Yu, F.~Chen, Physiological indicators for user trust
  in machine learning with influence enhanced fact-checking, in: Machine
  Learning and Knowledge Extraction: Third IFIP TC 5, TC 12, WG 8.4, WG 8.9, WG
  12.9 International Cross-Domain Conference, CD-MAKE 2019, Canterbury, UK,
  August 26--29, 2019, Proceedings 3, Springer, 2019, pp. 94--113.

\bibitem{vo2019learning}
N.~Vo, K.~Lee, Learning from fact-checkers: analysis and generation of
  fact-checking language, in: Proceedings of the 42nd international ACM SIGIR
  conference on research and development in information retrieval, 2019, pp.
  335--344.

\bibitem{krause2020fact}
N.~M. Krause, I.~Freiling, B.~Beets, D.~Brossard, Fact-checking as risk
  communication: the multi-layered risk of misinformation in times of covid-19,
  Journal of Risk Research 23~(7-8) (2020) 1052--1059.

\bibitem{abdeen2021fighting}
M.~A. Abdeen, A.~A. Hamed, X.~Wu, Fighting the covid-19 infodemic in news
  articles and false publications: The neonet text classifier, a supervised
  machine learning algorithm, Applied Sciences 11~(16) (2021) 7265.

\bibitem{siwakoti2021covid}
S.~Siwakoti, K.~Yadav, N.~Bariletto, L.~Zanotti, U.~Erdogdu, J.~N. Shapiro, How
  covid drove the evolution of fact-checking, Harvard Kennedy School
  Misinformation Review (2021).

\bibitem{koohi2023generative}
M.~Koohi-Moghadam, K.~T. Bae, Generative ai in medical imaging: applications,
  challenges, and ethics, Journal of Medical Systems 47~(1) (2023) 94.

\bibitem{degrave2023dissection}
A.~J. DeGrave, Z.~R. Cai, J.~D. Janizek, R.~Daneshjou, S.-I. Lee, Dissection of
  medical ai reasoning processes via physician and generative-ai collaboration,
  Medrxiv (2023).

\bibitem{trabassi2024optimizing}
D.~Trabassi, S.~F. Castiglia, F.~Bini, F.~Marinozzi, A.~Ajoudani, M.~Lorenzini,
  G.~Chini, T.~Varrecchia, A.~Ranavolo, R.~De~Icco, et~al., Optimizing rare
  disease gait classification through data balancing and generative ai:
  insights from hereditary cerebellar ataxia, Sensors 24~(11) (2024) 3613.

\bibitem{wang2023applications}
R.~Wang, V.~Bashyam, Z.~Yang, F.~Yu, V.~Tassopoulou, S.~S. Chintapalli,
  I.~Skampardoni, L.~P. Sreepada, D.~Sahoo, K.~Nikita, et~al., Applications of
  generative adversarial networks in neuroimaging and clinical neuroscience,
  Neuroimage 269 (2023) 119898.

\bibitem{tian2024opportunities}
S.~Tian, Q.~Jin, L.~Yeganova, P.-T. Lai, Q.~Zhu, X.~Chen, Y.~Yang, Q.~Chen,
  W.~Kim, D.~C. Comeau, et~al., Opportunities and challenges for chatgpt and
  large language models in biomedicine and health, Briefings in Bioinformatics
  25~(1) (2024) bbad493.

\bibitem{van2024chatgpt}
M.~M. Van~Wyk, Is chatgpt an opportunity or a threat? preventive strategies
  employed by academics related to a genai-based llm at a faculty of education,
  Journal of applied learning and teaching 7~(1) (2024) 35--45.

\bibitem{barreto2023generative}
F.~Barreto, L.~Moharkar, M.~Shirodkar, V.~Sarode, S.~Gonsalves, A.~Johns,
  Generative artificial intelligence: Opportunities and challenges of large
  language models, in: International conference on intelligent computing and
  networking, Springer, 2023, pp. 545--553.

\bibitem{giannakos2024promise}
M.~Giannakos, R.~Azevedo, P.~Brusilovsky, M.~Cukurova, Y.~Dimitriadis,
  D.~Hernandez-Leo, S.~J{\"a}rvel{\"a}, M.~Mavrikis, B.~Rienties, The promise
  and challenges of generative ai in education, Behaviour \& Information
  Technology (2024) 1--27.

\bibitem{Hamed_FC2004}
A.~A. Hamed, A.~Crimi, B.~S. Lee, M.~M. Misiak, Fact-checking generative ai:
  Ontology-driven biological graphs for disease-gene link verification, in:
  Computational Science – ICCS 2024: 24th International Conference, Malaga,
  Spain, July 2–4, 2024, Proceedings, Part IV, Springer-Verlag, Berlin,
  Heidelberg, 2024, p. 130–137.

\bibitem{augenstein2024factuality}
I.~Augenstein, T.~Baldwin, M.~Cha, T.~Chakraborty, G.~L. Ciampaglia, D.~Corney,
  R.~DiResta, E.~Ferrara, S.~Hale, A.~Halevy, et~al., Factuality challenges in
  the era of large language models and opportunities for fact-checking, Nature
  Machine Intelligence 6~(8) (2024) 852--863.

\bibitem{peng2023check}
B.~Peng, M.~Galley, P.~He, H.~Cheng, Y.~Xie, Y.~Hu, Q.~Huang, L.~Liden, Z.~Yu,
  W.~Chen, et~al., Check your facts and try again: Improving large language
  models with external knowledge and automated feedback, arXiv preprint
  arXiv:2302.12813 (2023).

\bibitem{mahmood2023fact}
R.~Mahmood, G.~Wang, M.~Kalra, P.~Yan, Fact-checking of ai-generated reports,
  in: International Workshop on Machine Learning in Medical Imaging, Springer,
  2023, pp. 214--223.

\bibitem{manakul2023selfcheckgpt}
P.~Manakul, A.~Liusie, M.~J. Gales, Selfcheckgpt: Zero-resource black-box
  hallucination detection for generative large language models, arXiv preprint
  arXiv:2303.08896 (2023).

\bibitem{yuan2022BioBART}
H.~Yuan, Z.~Yuan, R.~Gan, J.~Zhang, Y.~Xie, S.~Yu,
  \href{https://arxiv.org/abs/2204.03905}{Biobart: Pretraining and evaluation
  of a biomedical generative language model} (2022).
\newblock \href {http://arxiv.org/abs/2204.03905} {\path{arXiv:2204.03905}}.
\newline\urlprefix\url{https://arxiv.org/abs/2204.03905}

\bibitem{jin2023retrieve}
Q.~Jin, R.~Leaman, Z.~Lu, Retrieve, summarize, and verify: how will chatgpt
  affect information seeking from the medical literature?, Journal of the
  American Society of Nephrology 34~(8) (2023) 1302--1304.

\bibitem{hou2023answers}
Y.~Hou, J.~Yeung, H.~Xu, C.~Su, F.~Wang, R.~Zhang, From answers to insights:
  Unveiling the strengths and limitations of chatgpt and biomedical knowledge
  graphs, Research square (2023).

\bibitem{huly2024old}
O.~Huly, I.~Pogrebinsky, D.~Carmel, O.~Kurland, Y.~Maarek, Old ir methods meet
  rag, in: Proceedings of the 47th International ACM SIGIR Conference on
  Research and Development in Information Retrieval, 2024, pp. 2559--2563.

\bibitem{jeong2023generative}
C.~Jeong, Generative ai service implementation using llm application
  architecture: based on rag model and langchain framework, Journal of
  Intelligence and Information Systems 29~(4) (2023) 129--164.

\bibitem{arslan2024business}
M.~Arslan, S.~Munawar, C.~Cruz, Business insights using rag--llms: a review and
  case study, Journal of Decision Systems (2024) 1--30.

\bibitem{ng2024rag}
K.~K.~Y. Ng, I.~Matsuba, P.~C. Zhang, Rag in health care: A novel framework for
  improving communication and decision-making by addressing llm limitations,
  NEJM AI (2024) AIra2400380.

\bibitem{khan10803468MedAI}
Y.~Khan, A.~A. Hamed, { Reinforcement Explainability of ChatGPT Prompts by
  Embedding Breast Cancer Self-Screening Rules into AI Responses }, in: 2024
  IEEE International Conference on Medical Artificial Intelligence (MedAI), Los
  Alamitos, CA, USA, 2024, pp. 392--397.

\bibitem{Hamed10803434MedAI}
A.~A. Hamed, T.~E. Fandy, X.~Wu, { Accelerating Complex Disease Treatment
  Through Network Medicine and GenAI: A Case Study on Drug Repurposing for
  Breast Cancer }, in: 2024 IEEE International Conference on Medical Artificial
  Intelligence (MedAI), IEEE Computer Society, Los Alamitos, CA, USA, 2024, pp.
  354--359.

\bibitem{thomo2024pubmed}
A.~Thomo, Pubmed retrieval with rag techniques, in: Digital Health and
  Informatics Innovations for Sustainable Health Care Systems, IOS Press, 2024,
  pp. 652--653.

\bibitem{copilot2025}
{Microsoft}, {GitHub Copilot: AI-Powered Code Completion},
  \url{https://copilot.microsoft.com/}, accessed: 2025-01-12 (2025).

\bibitem{gemini2025}
{Google}, {Gemini: AI-Powered App by Google},
  \url{https://gemini.google.com/app}, accessed: 2025-01-12 (2025).

\bibitem{caramancion2023news}
K.~M. Caramancion, News verifiers showdown: a comparative performance
  evaluation of chatgpt 3.5, chatgpt 4.0, bing ai, and bard in news
  fact-checking, in: 2023 IEEE Future Networks World Forum (FNWF), IEEE, 2023,
  pp. 1--6.

\bibitem{singhal2024multilingual}
A.~Singhal, T.~Law, C.~Kassner, A.~Gupta, E.~Duan, A.~Damle, R.~Li,
  Multilingual fact-checking using llms, in: Proceedings of the Third Workshop
  on NLP for Positive Impact, 2024, pp. 13--31.

\bibitem{DeVerna_pnas_2322823121}
M.~R. DeVerna, H.~Y. Yan, K.-C. Yang, F.~Menczer, Fact-checking information
  from large language models can decrease headline discernment, Proceedings of
  the National Academy of Sciences 121~(50) (2024) e2322823121.
\newblock \href {https://doi.org/10.1073/pnas.2322823121}
  {\path{doi:10.1073/pnas.2322823121}}.

\bibitem{symp_ontology}
T.~S.~O. Developers, {Symptom Ontology (SYMP)},
  \url{http://purl.obolibrary.org/obo/symp.owl}, accessed: 2025-01-12 (2010).

\bibitem{schriml2022}
L.~M. Schriml, E.~Mitraka, J.~Munro, J.~C. Hu, V.~Felix, W.~A. Kibbe, {Human
  Disease Ontology 2022 update: improved classification, inferencing, and
  applications for genomics and disease research}, Nucleic Acids Research 50
  (2022) D1255--D1261.

\bibitem{schriml2012}
L.~M. Schriml, C.~Arze, S.~Nadendla, Y.~Chang, M.~Mazaitis, V.~Felix, G.~Feng,
  W.~A. Kibbe, {Disease Ontology: a backbone for disease semantic integration},
  Nucleic Acids Research 40 (2012) D940--D946.

\bibitem{chebi2006}
K.~Degtyarenko, P.~de~Matos, M.~Ennis, J.~Hastings, M.~Zbinden, A.~McNaught,
  R.~Alcantara, M.~Darsow, M.~Guedj, M.~Ashburner, {ChEBI}: a database and
  ontology for chemical entities of biological interest, Nucleic Acids Research
  36 (2008) D344--D350.

\bibitem{chebi2016}
J.~Hastings, G.~Owen, A.~Dekker, M.~Ennis, N.~Kale, V.~Muthukrishnan,
  S.~Turner, N.~Swainston, P.~Mendes, C.~Steinbeck, {ChEBI} in 2016: Improved
  services and an expanding collection of metabolites, Nucleic Acids Research
  44~(D1) (2016) D1214--D1219.

\bibitem{chebi2008}
J.~Hastings, P.~de~Matos, A.~Dekker, M.~Ennis, B.~Harsha, N.~Kale,
  V.~Muthukrishnan, G.~Owen, S.~Turner, M.~Williams, C.~Steinbeck,
  \href{https://www.ebi.ac.uk/chebi/}{The chebi reference database and ontology
  for biologically relevant chemistry: enhancements for 2013}, Nucleic Acids
  Research 41 (2013) D456--D463, accessed: 2025-01-12.
\newblock \href {https://doi.org/10.1093/nar/gks1146}
  {\path{doi:10.1093/nar/gks1146}}.
\newline\urlprefix\url{https://www.ebi.ac.uk/chebi/}

\bibitem{geneontology2000}
T.~G.~O. Consortium, \href{http://geneontology.org/}{Gene ontology: tool for
  the unification of biology}, Nature Genetics 25~(1) (2000) 25--29, accessed:
  2025-01-12.
\newblock \href {https://doi.org/10.1038/75556} {\path{doi:10.1038/75556}}.
\newline\urlprefix\url{http://geneontology.org/}

\bibitem{disease_ontology_doid10763}
D.~Ontology, Doid:10763 - lung cancer,
  \url{https://disease-ontology.org/term/DOID:10763/}, accessed: 2025-01-16
  (2025).

\bibitem{pubmed}
Pubmed, \url{https://pubmed.ncbi.nlm.nih.gov/}, accessed: February 4, 2025
  (n.d.).

\bibitem{jin2024pubmed}
Q.~Jin, R.~Leaman, Z.~Lu, Pubmed and beyond: biomedical literature search in
  the age of artificial intelligence, EBioMedicine 100 (2024) 104988, epub 2024
  Feb 1.
\newblock \href {https://doi.org/10.1016/j.ebiom.2024.104988}
  {\path{doi:10.1016/j.ebiom.2024.104988}}.

\bibitem{camon2004gene}
E.~Camon, M.~Magrane, D.~Barrell, V.~Lee, E.~Dimmer, J.~Maslen, D.~Binns,
  N.~Harte, R.~Lopez, R.~Apweiler, The gene ontology annotation (goa) database:
  sharing knowledge in uniprot with gene ontology, Nucleic acids research
  32~(suppl\_1) (2004) D262--D266.

\bibitem{yon2008use}
S.~Yon~Rhee, V.~Wood, K.~Dolinski, S.~Draghici, Use and misuse of the gene
  ontology annotations, Nature Reviews Genetics 9~(7) (2008) 509--515.

\bibitem{gene2012gene}
G.~O. Consortium, Gene ontology annotations and resources, Nucleic acids
  research 41~(D1) (2012) D530--D535.

\bibitem{drugbank_DB01259}
D.~Online, \href{https://go.drugbank.com/drugs/DB01259}{Lapatinib}, accessed:
  2023-01-04 (2023).
\newline\urlprefix\url{https://go.drugbank.com/drugs/DB01259}

\bibitem{Xu_2009}
F.~Xu, C.~Mao, Y.~Hu, C.~Rui, Z.~Xu, L.~Zhang, Cardiovascular effects of
  losartan and its relevant clinical application, Current Medicinal Chemistry
  16~(29) (2009) 3841--3857.
\newblock \href {https://doi.org/10.2174/092986709789178046}
  {\path{doi:10.2174/092986709789178046}}.

\bibitem{Suzuki_2015}
H.~Suzuki, K.~Shimada, K.~Fujiwara, Antihypertensive effectiveness of
  combination therapy with losartan/hydrochlorothiazide for 'real world'
  management of isolated systolic hypertension, Therapeutic Advances in
  Cardiovascular Disease 9~(1) (2015) 10--18, epub 2014 Nov 3.
\newblock \href {https://doi.org/10.1177/1753944714558244}
  {\path{doi:10.1177/1753944714558244}}.

\bibitem{microsoft2024phi3}
M.~AI, Phi-3: A family of small language models by microsoft, Online, available
  at: \url{https://azure.microsoft.com/en-us/products/phi} (2024).

\bibitem{tinyzero}
J.~Pan, J.~Zhang, X.~Wang, L.~Yuan, H.~Peng, A.~Suhr, Tinyzero,
  https://github.com/Jiayi-Pan/TinyZero, accessed: 2025-01-24 (2025).

\bibitem{meta2024llama}
M.~AI, Llama: Large language model meta ai, Online, available at:
  \url{https://www.llama.com/} (2024).

\bibitem{qwen2.5}
A.~Yang, B.~Yang, B.~Zhang, B.~Hui, B.~Zheng, B.~Yu, C.~Li, D.~Liu, F.~Huang,
  H.~Wei, H.~Lin, J.~Yang, J.~Tu, J.~Zhang, J.~Yang, J.~Yang, J.~Zhou, J.~Lin,
  K.~Dang, K.~Lu, K.~Bao, K.~Yang, L.~Yu, M.~Li, M.~Xue, P.~Zhang, Q.~Zhu,
  R.~Men, R.~Lin, T.~Li, T.~Xia, X.~Ren, X.~Ren, Y.~Fan, Y.~Su, Y.~Zhang,
  Y.~Wan, Y.~Liu, Z.~Cui, Z.~Zhang, Z.~Qiu, Qwen2.5 technical report, arXiv
  preprint arXiv:2412.15115 (2024).

\bibitem{qwen2}
A.~Yang, B.~Yang, B.~Hui, B.~Zheng, B.~Yu, C.~Zhou, C.~Li, C.~Li, D.~Liu,
  F.~Huang, G.~Dong, H.~Wei, H.~Lin, J.~Tang, J.~Wang, J.~Yang, J.~Tu,
  J.~Zhang, J.~Ma, J.~Xu, J.~Zhou, J.~Bai, J.~He, J.~Lin, K.~Dang, K.~Lu,
  K.~Chen, K.~Yang, M.~Li, M.~Xue, N.~Ni, P.~Zhang, P.~Wang, R.~Peng, R.~Men,
  R.~Gao, R.~Lin, S.~Wang, S.~Bai, S.~Tan, T.~Zhu, T.~Li, T.~Liu, W.~Ge,
  X.~Deng, X.~Zhou, X.~Ren, X.~Zhang, X.~Wei, X.~Ren, Y.~Fan, Y.~Yao, Y.~Zhang,
  Y.~Wan, Y.~Chu, Y.~Liu, Z.~Cui, Z.~Zhang, Z.~Fan, Qwen2 technical report,
  arXiv preprint arXiv:2407.10671 (2024).

\bibitem{khan2024reinforcement}
Y.~Khan, A.~A. Hamed, Reinforcement explainability of chatgpt prompts by
  embedding breast cancer self-screening rules into ai responses, in: 2024 IEEE
  International Conference on Medical Artificial Intelligence (MedAI), IEEE,
  2024, pp. 392--397.

\end{thebibliography}

\end{document}